%% file: main.tex
\documentclass[10pt,twocolumn,letterpaper]{article}

\usepackage[pagenumbers]{iccv} 
\definecolor{iccvblue}{rgb}{0.21,0.49,0.74}
\usepackage[pagebackref,breaklinks,colorlinks,allcolors=iccvblue]{hyperref}
\usepackage{amsmath}
\usepackage{mathrsfs}
\usepackage{graphicx}
\usepackage{booktabs}
\usepackage{array}
\usepackage{multirow}
\usepackage{pifont}
\usepackage{soul}
\usepackage[bottom]{footmisc}

\newcommand{\modelname}{BASIC}
\newcommand{\myparagraph}[1]{\textbf{#1}\hspace{1.8ex}}
\newcommand{\mysubparagraph}[1]{\textit{#1}\hspace{1.8ex}}

\def\paperID{15867} 
\def\confName{ICCV}
\def\confYear{2025}

\title{BASIC: Boosting Visual Alignment with Intrinsic Refined Embeddings in Multimodal Large Language Models}

\author{
Jianting Tang\textsuperscript{1,2}
\and
Yubo Wang\textsuperscript{1,2}
\and
Haoyu Cao\textsuperscript{1,2}
\and
Linli Xu\textsuperscript{1,2}\thanks{Corresponding author.}
\and
\textsuperscript{1}\text{University of Science and Technology of China}, 
\textsuperscript{2}\text{State Key Laboratory of Cognitive Intelligence}\\
{\tt\small \{jiantingtang,wyb123,caohaoyu\}@mail.ustc.edu.cn}, 
{\tt\small linlixu@ustc.edu.cn}
}

\begin{document}
\maketitle
\input{sec/0_abstract}
\input{sec/1_introduction}
\input{sec/2_related_work}
\input{sec/3_visual_analysis}
\input{sec/4_methodology}

\input{sec/5_experiments}

\input{sec/6_conclusion}
\input{sec/7_acknowledgements}
{
    \small
    \bibliographystyle{ieeenat_fullname}
    \bibliography{main}
}

\input{sec/X_suppl}
\end{document}

%% file: sec/0_abstract.tex
\begin{abstract}
Mainstream Multimodal Large Language Models (MLLMs) achieve visual understanding by using a vision projector to bridge well-pretrained vision encoders and large language models (LLMs). The inherent gap between visual and textual modalities makes the embeddings from the vision projector critical for visual comprehension.
However, current alignment approaches treat visual embeddings as contextual cues and merely apply auto-regressive supervision to textual outputs, neglecting the necessity of introducing equivalent direct visual supervision, which hinders the potential finer alignment of visual embeddings.
In this paper, based on our analysis of the refinement process of visual embeddings in the LLM's shallow layers, we propose \textbf{\modelname{}}, a method that utilizes refined visual embeddings within the LLM as supervision to directly guide the projector in generating initial visual embeddings.
Specifically, the guidance is conducted from two perspectives: ($i$) optimizing embedding directions by reducing angles between initial and supervisory embeddings in semantic space;
 ($ii$) improving semantic matching by minimizing disparities between the logit distributions of both visual embeddings.
Without additional supervisory models or artificial annotations, \modelname{} significantly improves the performance of MLLMs across a wide range of benchmarks, demonstrating the effectiveness of our introduced direct visual supervision.

\end{abstract}

%% file: sec/1_introduction.tex
\section{Introduction}
Multimodal Large Language Models (MLLMs)~\citep{internvl,qwen2vl,cogvlm,llava1.5,mplug} have demonstrated impressive performance on tasks requiring strong visual perception and logical reasoning, marking a solid step towards general artificial intelligence (AGI).
To efficiently construct high-performance MLLMs, a simple vision projector~\citep{llava1.5,llava1,Honeybee} is typically used to bridge well-pretrained powerful vision encoders~\citep{clip-vit,siglip,dinov2,sam} and large language models (LLMs)~\citep{llama3,qwen2,internlm,vicuna}.
To fully leverage both models, it is crucial to effectively align the visual and textual modalities through the vision projector~\citep{assignpred,blip2,mindgap}.

\begin{figure}[t!]
  \centering
  \includegraphics[width=\linewidth]{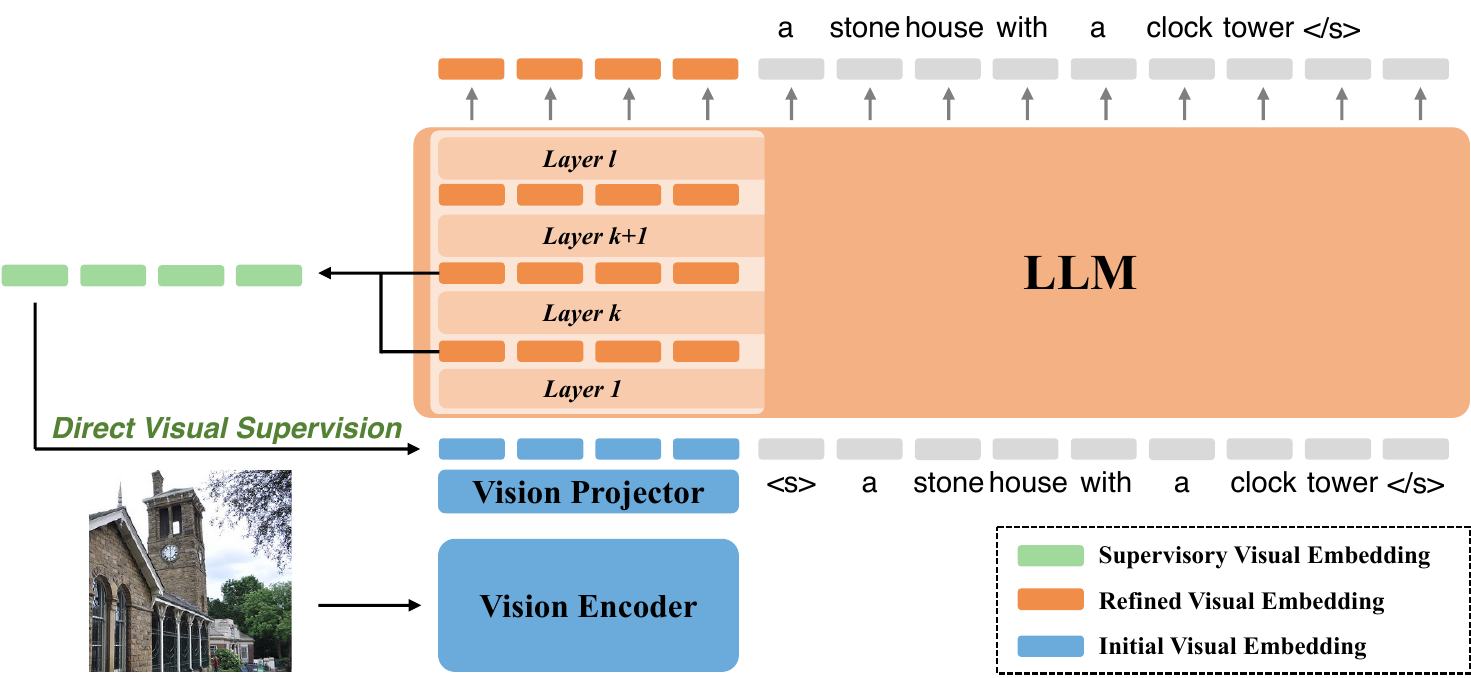}
  \vspace{-0.5cm}
  \caption{
  Overview of the proposed \modelname{} framework. Conventional MLLM training treats visual embeddings derived from the vision projector as contextual cues, only applying auto-regressive supervision to text tokens. Beyond that, \modelname{} leverages the refined visual embeddings from the LLM's shallow layers to provide direct visual supervision to the initial visual embeddings.
  }
  \label{fig:page_begin}
  \vspace{-0.5cm}
\end{figure}

Current leading MLLMs, such as LLaVA~\citep{llava1.5}, InternVL~\citep{internvl} and Qwen2-VL~\citep{qwen2vl}, employ a multistage training paradigm to achieve modality alignment.
Generally, the early stages establish basic multimodal understanding using large-scale image-caption pairs, aligning the vision projector's output with the LLM's input embedding space.
The subsequent stages refine this alignment by tuning on high-quality visual instruction-response data, developing the model's capacity for specific visual tasks.
However, due to the continuity of visual embeddings and the discreteness of text tokens, the current training approaches treat visual embeddings purely as contextual cues and apply auto-regressive supervision to  text tokens, which implies lack of equivalent direct supervision for visual embeddings. The asymmetric supervision adapted from the training paradigm targeted for LLMs leads to two key problems. First, it fails to fully utilize the rich information present in visual data~\citep{libra}. Second, it limits the model's ability to achieve fine-grained alignment between visual and linguistic representations~\citep{vwlmm}.

Recently, alternative works such as Chameleon~\citep{chameleon}, SEED-LLaMA~\citep{seed} and LaVIT~\citep{lavit} employ pre-trained image tokenizers to obtain discrete visual tokens, achieving unified auto-regressive modeling. However, while these approaches treat both modalities equally, the discretization process introduces significant information loss. 
Additionally, Emu1~\citep{emu1} and Emu2~\citep{emu2} use the $\ell_2$ regression loss to encourage each continuous visual representation output by the LLM to directly fit the input value at the next position.
Despite their impressive image generation ability benefiting from this unique design, they still lag behind mainstream MLLMs in visual comprehension.
To enhance visual comprehension by introducing direct visual supervision, it is essential to ensure the quality of the constructed supervisory signals and the rationality of the optimization objective during the training process.

In this paper, we first analyze the visual perception process in the well-established MLLMs. Based on our findings, we propose \modelname{}, a novel approach that leverages refined visual embeddings within LLMs as supervisory signals to boost modality alignment in the input space, as illustrated in Figure~\ref{fig:page_begin}. 

Our analysis begins by examining how visual embeddings relate to textual concepts. For each initial visual embedding derived from the vision projector, we calculate its cosine similarity with all text token embeddings in the LLM's vocabulary and visualize the most matching token.
As shown in Figure~\ref{fig:visualize}, for certain visual embeddings, the most matching text tokens directly reflect attributes of corresponding image patches, such as color, shape, and object class. 
This partially reveals the internal visual perception mechanism of MLLMs: LLMs interpret the textual concepts within visual embeddings to understand images. However, there are still plenty of initial visual embeddings associated with irregular text tokens. Tracking these embeddings through the LLM's layers, we observe that, in shallow layers, initially misaligned embeddings often gradually align with more meaningful text tokens, which should be attributed to the LLM's strong semantic modeling capabilities by considering the visual context. 
Despite the gradual refinement of visual embeddings within the LLM, the visual embeddings seen by the \textit{questions} in the early stages are inaccurate, which can lead to confusion in image understanding and impair the final \textit{answers}. Therefore, it is crucial that initial visual embeddings exhibit high quality from the outset.

To this end, we construct direct visual supervision by weighted summation of refined visual embeddings from the shallow layers. 
Firstly, we optimize the directional alignment between initial and supervisory visual embeddings by minimizing their angular distances in semantic space. 
Secondly, to ensure the consistency of semantic distribution, we compute logit distributions across the entire vocabulary for both initial and supervisory visual embeddings, and then minimize their KL divergence. 
Our approach ensures the visual embeddings maintain consistent high quality in the LLM's shallow layers, enabling the \textit{questions} to acquire accurate image information at an early stage.

Notably, our method exhibits the following advantages. Firstly, it does not rely on additional supervisory models or artificial annotations, thus saving the resource overhead. Furthermore, it is generally applicable to a wide range of MLLMs adopting the vision encoder-vision projector-LLM architecture. Comprehensive experiments demonstrate the effectiveness of our introduced direct visual supervision.

In summary, our contributions are as follows:
\begin{itemize}

\item We systematically analyze the association between the visual embeddings within different LLM layers and the text token embeddings, which provides valuable insights into the internal visual perception mechanism of MLLMs.

\item We propose \modelname{}, an effective direct visual supervision method that leverages the LLM's internal refined visual embeddings to guide initial visual embeddings in the input space from two perspectives: the directional alignment and semantic distribution. 

\item \modelname{} notably improves the performance of a series of MLLMs across a wide range of benchmarks, demonstrating its applicability and robustness.

\end{itemize}

%% file: sec/2_related_work.tex
\section{Related Work}

\begin{figure*}[t!]
  \centering
  \includegraphics[width=\linewidth]{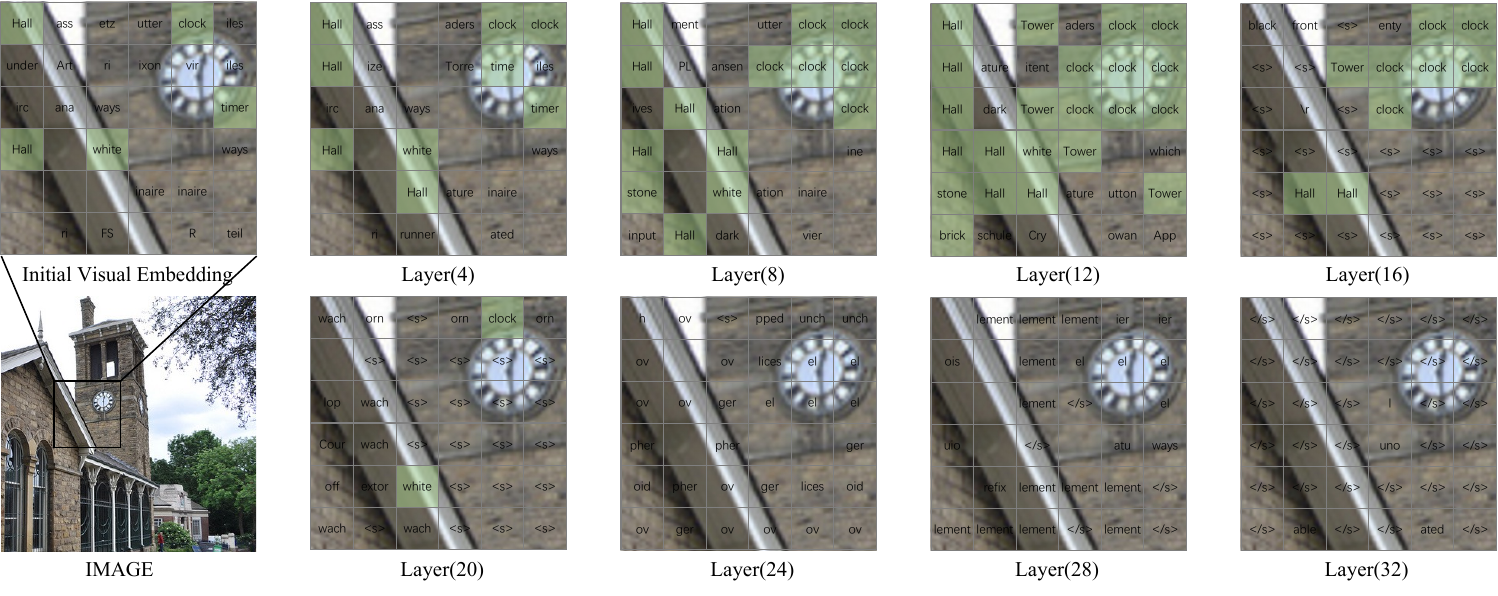}
  \vspace{-0.6cm}
  \caption{
    Visualization of the closest matching text token for each visual embedding across different layers of the LLM. Green patches indicate semantically meaningful matches.
    The initial visual embeddings are derived from the vision projector and have not yet entered the LLM. Layer($\cdot$) indicates the matching results of the visual embeddings from the corresponding LLM layer. There are two notable patterns: $(i)$ in the shallow layers of the LLM, visual embeddings that initially correspond to irregular tokens gradually align with more meaningful tokens; $(ii)$ in the deep layers of the LLM, visual embeddings tend to correspond with the special end token \texttt{</s>}.
  }
  \label{fig:visualize}
  \vspace{-0.3cm}
\end{figure*}

\subsection{Multimodal Large Language Models}
At present, most prevailing MLLMs, such as LLaVA~\citep{llava1.5}, InternVL~\citep{internvl} and Qwen2-VL~\citep{qwen2vl}, adopt a vision encoder-vision projector-LLM architecture. The vision projector~\citep{llava1,blip2,flamingo,perceiver} is responsible for mapping the image features encoded by the vision encoders into the LLM's input embedding space.
This simple approach broadens the LLM's comprehension to images.
However, the approach of independently training and then grafting inherently leads to the difficulty of modality alignment.
Recently, another technical route~\citep{chameleon,seed,lavit} employs an image tokenizer~\citep{vqvae,beit,beit2,magvit2} to obtain discrete visual tokens and conducts unified auto-regressive modeling to create native MLLMs. However, it requires extensive training to converge, and the discretization process leads to substantial visual information loss.
Therefore, this paper follows the first technical route, and utilize refined visual embeddings within the LLM to construct additional direct visual supervision to boost modality alignment.

\subsection{Mechanistic Interpretability}
Understanding the internal visual perception mechanism of MLLMs is crucial for constructing effective direct visual supervision.
Mechanistic interpretility~\citep{interpret1,interpret2,interpret3} aims to 
uncover the internal mechanisms that drive the input-output transformation.
Sparse autoencoder(SAE) based methods~\citep{sae1,sae2,sae3,sae4} utilize the representation reconstruction and sparsification to facilitate discovering semantic features in sparse representations.
Logit lens based methods~\citep{logitlens1,logitlens2,logitlens3,logitlens4} use the language model head to project hidden states to interpret the prediction process. Currently, most interpretability studies mainly focus on LLMs, which only involves the text modality. The systematic analysis of visual perception process within MLLMs remains a rather unexplored field. 

\subsection{Self-Distillation}
Self-distillation~\citep{selfdistill1,selfdistill2,selfdistill3} is a unique instance of knowledge distillation~\citep{kd1,kd2,kd3,kd4} where the teacher and student models share the same architecture. The network leverages its internally learned knowledge to guide its own training. Some works~\citep{lastbatch1,lastbatch2} employ models updated at earlier steps as teachers for the current step, facilitating knowledge transfer across the temporal dimension. Some studies~\citep{byot,sd1,sd2,sssd} divide the model into different parts and use deeper blocks as teachers for shallower blocks, achieving knowledge transfer within the spatial dimension. 
From this perspective, utilizing the refined visual embeddings within LLM's shallow layers to guide the vision projector in generating better-aligned initial visual embeddings can be regarded as a form of self-distillation.
\vspace{-6.8pt}

%% file: sec/3_visual_analysis.tex
\section{Visual Perception Process Analysis}
\subsection{Preliminary}
MLLMs typically comprise three components: a vision encoder $F_v(\cdot)$, a vision projector $F_p(\cdot)$, and an LLM $F_t(\cdot)$.
Specifically, the vision encoder $F_v(\cdot)$ extracts visual features from raw images.
The vision projector $F_p(\cdot)$ transforms these image features into initial visual embeddings $V \in \mathbb{R}^{m \times d}$, where $m$ denotes the number of image patches and $d$ is the embedding dimension. 
The textual inputs are tokenized into token ids, which are then used to retrieve the corresponding token embeddings from the LLM's embedding layer $E \in \mathbb{R}^{N \times d}$, where $N$ represents the vocabulary size. The resulting textual embeddings are denoted as $T \in \mathbb{R}^{n \times d}$, where $n$ denotes the number of text tokens.
The LLM $F_t(\cdot)$ is responsible for processing the initial visual embeddings $V$ and textual embeddings $T$ to generate the final output. 
This can be formalized as:
\begin{equation}
\begin{aligned}
V = F_p(F_v(\text{image}&)), \quad T = E(\text{text}); \\
\text{output} &= F_t(V, T).
\end{aligned} 
\end{equation}

In the forward propagation process, the LLM gradually refines the initial visual embeddings $V$, and the resulting hidden states are referred as refined visual embeddings $\tilde{V} \in \mathbb{R}^{l \times m \times d}$, where $l$ represents the number of LLM layers. 

\begin{figure}[t!]
  \centering
  \includegraphics[width=\linewidth]{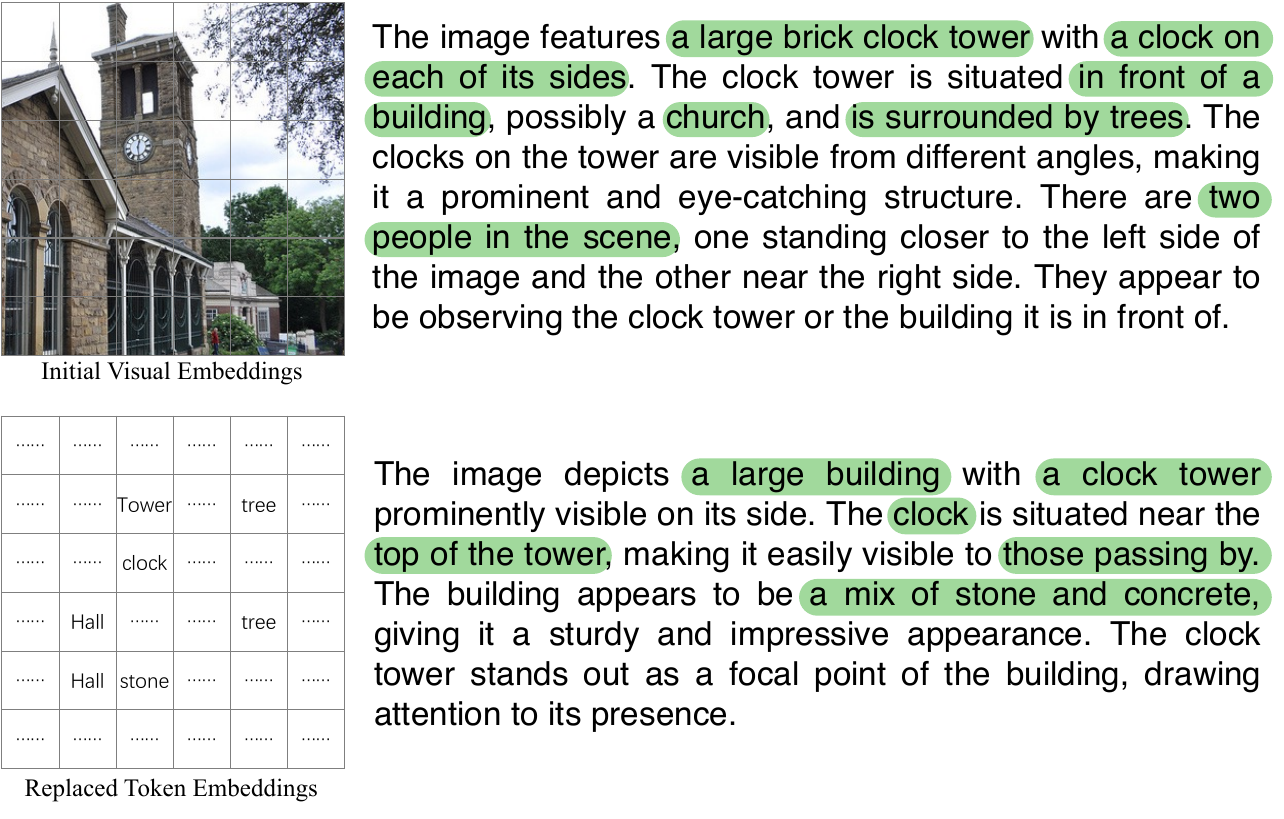}
  \vspace{-0.7cm}
  \caption{
    Visual embeddings from the vision projector are replaced with their closest matching text token embeddings in the LLM's vocabulary. The model is then prompted to generate descriptions. The generated descriptions (bottom) demonstrate strong semantic consistency with the raw image content.
  }
  \vspace{-0.4cm}
  \label{fig:replace}
\end{figure}

\subsection{Similarity-Based Analysis}
To construct effective direct supervision for each visual embedding, we first analyze the visual perception process within MLLMs. 
For each initial visual embedding $\boldsymbol{v}_i$ derived from the vision projector, the most semantically matching text token embedding $\boldsymbol{e}_i$ can be obtained as:
\begin{equation}
\boldsymbol{e}_i = \arg\max_j \frac{\boldsymbol{v}_i \cdot \boldsymbol{e}_j}{\|\boldsymbol{v}_i\|_2\|\boldsymbol{e}_j\|_2}
\end{equation}
where $0 \leq j < N$, $\boldsymbol{v}_i \in V$ and $\boldsymbol{e}_j \in E$.  
 As illustrated in Figure~\ref{fig:visualize} (upper left), for certain initial visual embeddings, the most similar tokens directly reflect specific attributes of the corresponding image patches, such as \textit{clock} expressing the class of object, and \textit{white} reflecting the color of object. 
Accordingly, we replace initial visual embeddings $\boldsymbol{[}\boldsymbol{v}_1, \boldsymbol{v}_2, \dots, \boldsymbol{v}_m\boldsymbol{]}$ with the most matched token embeddings $\boldsymbol{[}\boldsymbol{e}_1, \boldsymbol{e}_2, \dots, \boldsymbol{e}_m\boldsymbol{]}$, and prompt the LLM to generate image descriptions based on these replaced token embeddings. As shown in Figure~\ref{fig:replace}, the descriptions are highly consistent with the content of the raw image.
This indicates that LLMs might interpret the textual concepts within visual embeddings to understand images, and it also inspires us that direct visual supervision should guide the initial visual embeddings to establish more accurate text token associations.

Furthermore, to fully reveal the processing path of visual information within the LLM, we apply the same operations on all refined visual embeddings from different layers of the LLM.
As illustrated in Figure~\ref{fig:visualize}, there are significant pattern differences between the LLM's shallow and deep layers. In the LLM's shallow layers, plenty of visual embeddings that initially correspond to irregular tokens will gradually align with more meaningful tokens. In the LLM's deep layers, visual embeddings tend to correspond with the special end token \texttt{</s>}. 
Existing interpretability analyses targeted for LLMs~\citep{logitlens1,logitlens2,logitlens3} have shown that, when processing text, the shallow layers mainly focus on constructing better textual semantic representations by considering the context, while the deep layers concentrate on predicting the next token. 
In terms of visual embeddings, this enhanced semantic representations in the LLM's shallow layers offer an intuitive presentation, namely a better relevance of associated text tokens.
Meanwhile, due to the lack of token-level labels for visual inputs, the LLM's deep layers tend to predict \texttt{</s>} for each visual embedding to terminate the output.

Despite the refinement of visual embeddings in the shallow layers, the \textit{questions} perceive plenty of inaccurate visual embeddings in the early stages, which would cause significant confusion in image understanding.
Therefore, we leverage the refined visual embeddings from the LLM's shallow layers to guide the initial visual embeddings, boosting vision-language alignment from the early stage.

%% file: sec/4_methodology.tex
\begin{figure*}[t!]
  \centering
  \includegraphics[width=\linewidth]{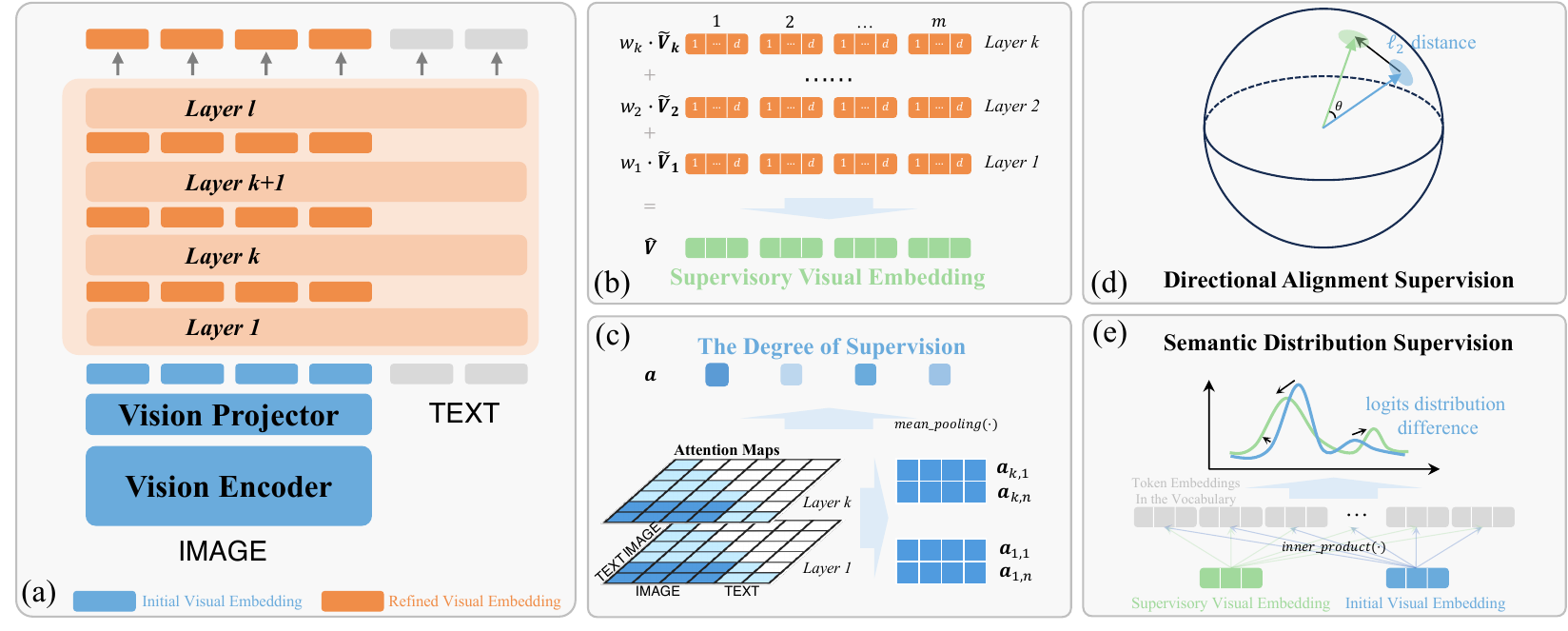}
  \vspace{-0.7cm}
  \caption{
    The construction of supervisory visual embeddings and two optimization objectives. (a) \textbf{Embedding Refinement: }The LLM processes the initial visual embeddings derived from the vision projector, generating refined visual embeddings at each layer. (b) \textbf{Supervisory Signal Construction:} Refined visual embeddings $\tilde{V}_i$ from layer $1$ to $k$ are weighted averaged by $w_i$ to serve as the supervisory visual embedding. (c) \textbf{Attention-based Supervision Degree:} Attention scores from text tokens to visual embeddings are mean-pooled to serve as the degree of supervision on each visual embedding. 
    (d) \textbf{Directional Alignment Supervision:} Initial and supervisory embeddings are aligned by narrowing the angle $\theta$ on the unit hypersphere.
    (e) \textbf{Semantic Distribution Supervision:} Logit distributions are computed by projecting both initial and supervisory embeddings against the LLM's vocabulary, then aligned by the minimizing KL divergence .
  }
  \label{fig:framework}
  \vspace{-0.3cm}
\end{figure*}

\section{Methodology}
In the general training process of MLLMs, the auto-regressive supervision is performed on all input text tokens, and the training loss can be expressed as:
\begin{align}
\mathcal{L}_{lm} = - \sum_{i=1}^{n} \log p(\boldsymbol{t}_i \mid V, \boldsymbol{t}_{<i})
\end{align}

\subsection{Supervisory Visual Embedding}
This section details the construction of supervisory visual embeddings.
In order to comprehensively utilize refined visual embeddings from the LLM's shallow layers to provide robust direct visual supervision, we assign weight $w_i$ for each LLM layer and 
obtain the final supervisory visual embedding $\hat{V}$:
\begin{align}
\hat{V} = \sum_{i=1}^{k} w_i \tilde{V}_i; \quad w_i = \frac{i^2}{\sum_{i=1}^{k} i^2}
\end{align}
where $\tilde{V}_i \in \mathbb{R}^{m \times d}$, $\hat{V} \in \mathbb{R}^{m \times d}$, $\sum w_i = 1$ and we use the refined visual embeddings from layer $1\!\sim\!k$. Considering the refining effect of LLM on visual embeddings, $w_i$ is set to increase quadratically with the layer number $i$. 

In each data sample, image patches contribute unequally to the model's text outputs, with more important image patches having a greater impact on the outputs. 
Therefore, we apply stronger supervisions to the initial visual embeddings from more important patches. Specifically, we use the attention scores from the text part across the LLM's layers to measure the importance of $i$th image patch:
\begin{align}
a_i = \frac{1}{kn} \sum_{h=1}^{k} \sum_{j=1}^{n} a_{h,j,i}; \quad a_i = \frac{a_i}{\sum_{i=1}^{m} a_i}
\end{align}
where $a_{h,j,i}$ denotes the attention score from the $j$th text token to $i$th image patch in the LLM's $h$th layer. 
Finally, $a_i$ denotes the mean attention score on the $i$th image patch and serves as the degree of supervision.

\subsection{Directional Alignment Supervision}
In the embedding space of LLMs, semantically similar embeddings could be measured by the cosine similarity\citep{logitlens1,logitlens2}. The cosine similarity essentially characterizes the angular relationship between embeddings. Therefore, we guide the initial visual embeddings $V$ to align with the direction of supervisory visual embeddings $\hat{V}$ to improve the semantic of $V$.
Specifically, we first normalize the embeddings to eliminate magnitude effects and then reduce the angle by minimizing their $\ell_2$ distance on the unit hypersphere:
\begin{align}
\mathcal{L}_{das} = \sum_{i=1}^{m} a_i \left\| \frac{\boldsymbol{v}_i}{\|\boldsymbol{v}_i\|_2} - \frac{\boldsymbol{\hat{v}}_i}{\|\boldsymbol{\hat{v}}_i\|_2} \right\|_2^2
\end{align}
where $\boldsymbol{v}_i \in \mathbb{R}^d$, $\boldsymbol{\hat{v}}_i \in \mathbb{R}^d$, and $a_i$ is used to control the degree of supervision.

\subsection{Semantic Distribution Supervision}
To analyze the association between visual and token embeddings, we compute the inner product of the visual embedding with the entire LLM vocabulary, yielding an interpretable logits vector. Each dimension of this logits vector reflects the semantic association between the visual embedding and the corresponding token, with higher values indicating tighter associations.
The logits vector comprehensively characterizes the global semantic distribution of each visual embedding on the whole vocabulary. 
Therefore, we utilize the supervisory visual embeddings to guide the initial visual embeddings to learn better textual associations. Specifically, we first compute the logits vectors $P$ between the supervisory visual embedding $\hat{V}$ and token embeddings $E$, as well as the logits vectors $Q$ between the initial visual embeddings $V$ and token embeddings $E$. We then minimize the KL divergence between both logits vectors to match the semantic distribution:

\begin{equation} 
\begin{aligned}
P = \text{softm}&\text{ax}(\hat{V}E^{\top}), \quad Q = \text{softmax}(VE^{\top}) \\
\mathcal{L}_{sds} &= \sum_{i=1}^{m} a_i \text{KL}\left( \boldsymbol{p}_i||\boldsymbol{q}_i \right)
\end{aligned}
\end{equation}
where $P, Q \in \mathbb{R}^{m \times N}$, $\boldsymbol{p_i},\boldsymbol{q_i} \in \mathbb{R}^{N}$, $m$ represents the number of image patches and $N$ represents the vocabulary size.
Overall, the total training loss used in the pre-training stage and instruction-tuning stage is:
\begin{align}
\mathcal{L} = \mathcal{L}_{lm} + \lambda_1 \mathcal{L}_{das} + \lambda_2 \mathcal{L}_{sds}
\end{align}
where $\lambda_1$ and $\lambda_2$ are used to balance different losses.

%% file: sec/5_experiments.tex


\begin{table*}[t!]
    \small
    \centering
    \resizebox{\textwidth}{!}{
    \begin{tabular}{l|cc|cccccccc}
        \toprule
        Method & LLM & Res. & $\text{VQA}^\text{v2}$ & GQA & $\text{SQA}^\text{I}$  & $\text{VQA}^\text{T}$ & $\text{MMB}^\text{EN}$ & $\text{MMB}^\text{CN}$ & MM-Vet & VizWiz  \\
        
        \hline\hline
        \rowcolor[gray]{0.85}\multicolumn{11}{l}{\textit{\textbf{Models using 7B LLM}}} \\ 
        \hline
        \addlinespace[0.5ex]
        Fuyu~\citep{fuyu} & Fuyu-8B & - & 74.2 & - & - & - & 10.7 & - & 21.4 & 35.9 \\
        LaVIT-v2~\cite{lavit} & LLaMA2-7B & 224 & 68.2 & 48.0 & - & - & - & - & - & 41.0 \\
        IDEFICS~\citep{idefics}       & LLaMA-7B & 224 & 50.9 & 38.4 & - & 25.9 & 48.2 & 25.2 & - & 35.5 \\
        InstructBLIP~\citep{instructblip} & Vicuna-7B & 224 & - & 49.2 & 60.5 & 50.1 & 36.0 & 23.7 & 26.2 & 34.5 \\
        Qwen-VL-Chat~\citep{qwenvl}         & Qwen-7B & 224 & 78.2 & 57.5 & 68.2 & 61.5 & 60.6 & 56.7 & - & 38.9 \\
        VW-LMM~\citep{vwlmm}           & Vicuna-7B & 336 & 78.9 & 62.7 & 68.1 & 57.6 & 65.9 & 59.8 & 31.3 & 48.3 \\
        LLaVA-1.5~\citep{llava1.5}   & Vicuna-7B & 336 & 78.5 & 62.0 & 66.8 & 58.2 & 64.3 & 58.3 & 31.1 & 50.0 \\
        \textbf{\modelname{}}   & Vicuna-7B & 336 & $\textbf{79.2}_{{\uparrow0.7}}$ & $\textbf{63.5}_{{\uparrow1.5}}$ & $\textbf{70.6}_{{\uparrow3.8}}$ & $58.0_{\textcolor{gray}{\downarrow0.2}}$ & $\textbf{68.8}_{{\uparrow4.5}}$ & $\textbf{62.1}_{{\uparrow3.8}}$ & $\textbf{33.8}_{{\uparrow2.7}}$ & $\textbf{52.5}_{{\uparrow2.5}}$ \\
        
        \midrule\hline
        \rowcolor[gray]{0.85}\multicolumn{11}{l}{\textit{\textbf{Models using 13B LLM}}} \\ 
        \hline
        \addlinespace[0.5ex]
        Emu-I~\citep{emu1}       & LLaMA-13B & 224 & 62.0 & 46.0 & - & - & - & - & 36.3 & 38.3 \\
        BLIP-2~\citep{blip2}    & Vicuna-13B & 224 & 65.0 & 41 & 61 & 42.5 & - & - & 22.4 & 19.6 \\
        InstructBLIP~\citep{instructblip}       & Vicuna-13B & 224 & - & 49.5 & 63.1 & 50.7 & - & - & 25.6 &  33.4 \\
        LLaVA-1.5~\citep{llava1.5}   & Vicuna-13B & 336 & 80.0 & 63.3 & 71.6 & 61.3 & 67.7 & 63.6 & 36.1 & 53.6 \\
        \textbf{\modelname{}}   & Vicuna-13B & 336 & $\textbf{80.6}_{{\uparrow0.6}}$ & $\textbf{64.6}_{{\uparrow1.3}}$ & $\textbf{73.1}_{{\uparrow1.5}}$ & $61.0_{\textcolor{gray}{\downarrow0.3}}$ & $\textbf{69.6}_{{\uparrow1.9}}$ & $\textbf{64.9}_{{\uparrow1.3}}$ & $\textbf{37.2}_{{\uparrow1.1}}$ & $\textbf{55.8}_{{\uparrow2.2}}$ \\
        \bottomrule
    \end{tabular}
    }
    \vspace{-0.2cm}
    \caption{Comparison with leading representative MLLMs on 8 popular benchmarks. Res. represents the image resolution of vision encoder. The \textbf{best results} are highlighted. Fuyu~\citep{fuyu} discards the vision encoder and merely relies on the LLM to model both images and text. LaVIT-v2~\citep{lavit} discretizes the image using a pre-trained image tokenizer. Emu1~\citep{emu1} drives each continuous visual representation outputted by the LLM to fit the input value at the next position. 
    }
    \label{tab:main_result}
    \vspace{-0.4cm}
\end{table*}

\section{Experiments}
\subsection{Experimental Setup}
\noindent  \myparagraph{Model Settings.}
In main experiments, we adopt CLIP-ViT-L/14-336px~\citep{clip-vit} as the vision encoder, a two-layer MLP with GeLU activation function as the vision projector and Vicuna-v1.5~\citep{vicuna} as the LLM.

\noindent  \myparagraph{Data and Training Details.}
In the pre-training stage, only the vision projector is trainable. The dataset used is LLaVA-1.5-558k~\citep{llava1.5}, composed of image and caption pairs.
The training epoch is 1, with a batch size of 256.
We utilize the weighted summation of refined visual embeddings from the LLM's shallow layers (specifically, layers $1\!\sim\!16/32$ of the Vicuna-v1.5-7B and layers $1\!\sim\!20/40$ of the Vicuna-v1.5-13B) as the supervisory visual embedding, with a quadratically increasing weighting coefficient $w_i$. 
The loss coefficients $\lambda_1$ and $\lambda_2$ are set to 1 and 0.01 respectively.
The learning rate is 1e-3.

In the instruction-tuning stage, both the vision projector and LLM are trainable. The multi-modal instruction dataset adopted is LLaVA-1.5-mix-665k~\citep{llava1.5}, comprising visual instruction-response pairs from various sources including VQAv2~\citep{vqav2}, ShareGPT~\citep{sharegpt}, RefCOCO\citep{refcoco1,refcoco2} and others. We finetune \modelname{} for 1 epoch with a batch size of 128. The learning rate is 2e-5. All other settings remain the same as in the previous stage.

\noindent  \myparagraph{Evaluations.}
To thoroughly assess the effectiveness of our method, we conduct evaluations across a wide range of benchmarks. This includes four popular general VQA benchmarks: VQA-v2~\citep{vqav2}, GQA~\citep{gqa}, ScienceQA-IMG~\citep{sqai} and TextVQA~\citep{vqat}. In addition, we adopt four benchmarks specifically targeting MLLMs and involving more comprehensive ability assessments: MMBench~\citep{mmbench}, MMBench-CN~\citep{mmbench}, MM-Vet~\citep{mmvet} and VisWiz~\citep{vizwiz}. 

\begin{figure*}[t!]
  \centering
  \includegraphics[width=\linewidth]{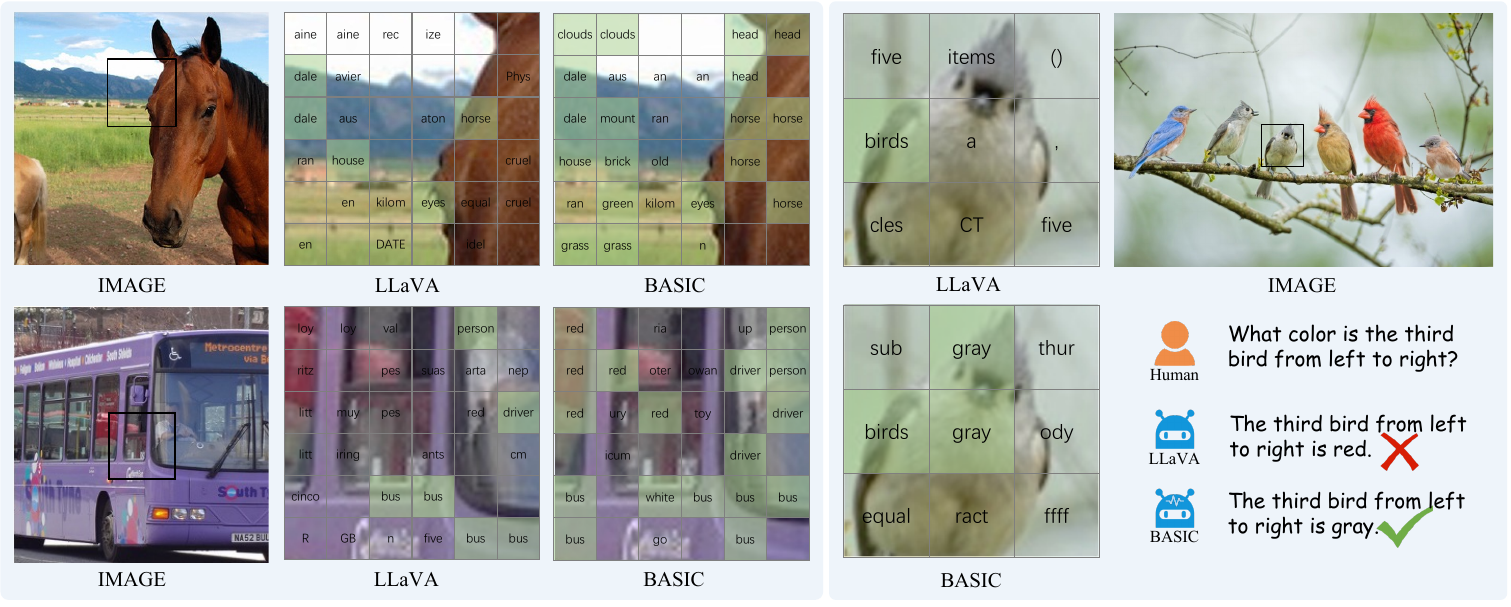}
  \vspace{-0.4cm}
  \caption{
   Comparisons between LLaVA~\citep{llava1.5} and \modelname{} in comprehending images. (Left) The respective closest matching text token for each initial visual embedding. (Right) More reasonable textual associations help \modelname{} solve visual comprehension tasks.
  }
  \label{fig:expvisual}
\end{figure*}

\begin{table*}[t!]
    \small
    \centering
    \resizebox{\textwidth}{!}{
    \begin{tabular}{l|cc|cccccccc}
        \toprule
        Method & $\mathcal{L}_{das}$ & $\mathcal{L}_{sds}$ & $\text{VQA}^\text{v2}$ & GQA & $\text{SQA}^\text{I}$  & $\text{VQA}^\text{T}$ & $\text{MMB}^\text{EN}$ & $\text{MMB}^\text{CN}$ & MM-Vet & VizWiz  \\
        
        \hline        
        \multirow{4}{*}{\modelname{}-7B} & \ding{55} & \ding{55} & 78.5 & 62.0 & 66.8 & \textbf{58.2} & 64.3 & 58.3 & 31.1 & 50.0 \\
        & \ding{51} & \ding{55} & 78.9 & 63.0 & \underline{68.5} & 57.7 & \underline{68.6} & \underline{61.4} & \underline{33.1} & \underline{51.4} \\
        & \ding{55} & \ding{51} & \underline{79.1} & \underline{63.3} & 68.1 & 57.6 & 68.0 & 60.6 & 32.5 & 51.2 \\
        & \ding{51} & \ding{51} & \textbf{79.2} & \textbf{63.5} & \textbf{70.6} & \underline{58.0} & \textbf{68.8} & \textbf{62.1} & \textbf{33.8} & \textbf{52.5} \\
        \bottomrule
    \end{tabular}
    }
    \vspace{-0.2cm}
    \caption{
    Contributions of each supervisory loss to the performance of MLLM. Results demonstrate the effectiveness of both losses.
    }
    \label{tab:two_losses}
    \vspace{-0.3cm}
\end{table*}

\subsection{Main Results}
As shown in Table~\ref{tab:main_result}, \modelname{} achieves superior results on various benchmarks.
Notably, \modelname{} utilizes the same model settings and training data as LLaVA-1.5~\citep{llava1.5}, the obviously improved performance highlighting the effectiveness of our method in promoting modality alignment and this basic factor benefiting a wide range of benchmarks.
Additionally, we notice that \modelname{}'s performance on $\text{VQA}^\text{T}$~\citep{vqat} slightly declines. This benchmark primarily assesses the model's textual content recognizing ability, with questions generally like \textit{``What is the year on the calendar?''} and \textit{``What’s the letter next to the z on the machine?''}. 
As the supervisory signal from the LLM’s shallow layers focuses more on “semantic concepts”, it might blur the abstract textual information within initial visual embeddings and affect scenarios with tiny text. 
To understand the reason behind the performance differences between \modelname{} and LLaVA~\citep{llava1.5}, we visualize the closest matching text token for each initial visual embedding, respectively. As illustrated in Figure~\ref{fig:expvisual}, the initial visual embeddings generated by \modelname{} have improved semantic and associate with more meaningful tokens. 
For an quantitative evaluation, we randomly select 30 images from the MS COCO dataset and have 2 master students count the number of “meaningful” initial visual embeddings in LLaVA and \modelname{}, respectively. The results show that \modelname{} outperforms in 100\% images, increasing the average ratio of meaningful embeddings from 74/576 to 217/576. The improved initial visual embeddings ensure the consistent high quality of visual embeddings in the LLM's shallow layers, allowing the \textit{questions} perceive accurate visual embeddings from the outset and reducing the confusion in image understanding.

Fuyu~\citep{fuyu} completely discards the well-trained vision encoder and directly uses a simple linear layer to transform image patches into input embeddings, entirely relying on the LLM to model both images and text. LaVIT-v2~\citep{lavit} employs an image tokenizer to encode images into discrete visual tokens, thus enabling unified auto-regressive modeling. These works represent initial efforts to develop native MLLMs, but they still significantly lag behind those adopting a vision encoder-vision projector-LLM architecture in image comprehension. VW-LLM~\citep{vwlmm} introduces an additional VM-head on the basis of LLaVA to provide auto-regressive supervision for the continuous image features outputted by the LLM. However, the requirement for specialized training for the VM-head finally resulting in a complex four-stage training pipeline.

\subsection{Ablation Studies}
\noindent  \myparagraph{Analysis of Supervisory Losses.}
To assess the contributions of two proposed optimization objectives respectively, we compare the performance of models obtained under three different training settings: using only $\mathcal{L}_{das}$, only $\mathcal{L}_{sds}$, and a combination of both.
The experiments are conducted on \modelname{}-7B, composed of CLIP-ViT-L/14-336px~\citep{clip-vit} and Vicuna-7B~\citep{vicuna}. The training process is consistent with the main experiment,
first pre-training on the LLaVA 1.5-558k dataset, and then instruction-tuning on the LLaVA 1.5-mix-665k dataset. 
As shown in Table~\ref{tab:two_losses}, when $\mathcal{L}_{das}$ and $\mathcal{L}_{sds}$ are not used, \modelname{} represents the original LLaVA~\citep{llava1.5}. The additional introduction of either $\mathcal{L}_{das}$ or $\mathcal{L}_{sds}$ can effectively enhance the model's performance across various benchmarks, demonstrating the effectiveness of guiding initial visual embeddings using supervisory visual embeddings from two distinct perspectives: the directional alignment and the semantic distribution. Additionally, the simultaneous utilization of both losses can further enhance the model's performance.

\begin{table*}[t!]
    \small
    \centering
    \resizebox{\textwidth}{!}{
    \begin{tabular}{l|ccc|cccccccc}
        \toprule
        Method & VE & LLM & PT+IT & $\text{VQA}^\text{v2}$ & GQA & $\text{SQA}^\text{I}$  & $\text{VQA}^\text{T}$ & $\text{MMB}^\text{EN}$ & $\text{MMB}^\text{CN}$ & MM-Vet & VizWiz  \\

        \hline
        LLaVA & \multirow{2}{*}{CLIP-L~\citep{clip-vit}} & \multirow{2}{*}{Gemma-2B~\citep{gemma}} & \multirow{2}{*}{558K+665K} & 72.5 & 56.0 & 61.3 & 43.7 & 54.0 & 49.5 & 23.9 & 38.7 \\
        \modelname{} & & & & \textbf{73.1} & \textbf{56.8} & \textbf{62.5} & \textbf{43.7} & \textbf{55.8} & \textbf{51.2} & \textbf{24.6} & \textbf{40.4} \\
        \midrule

        LLaVA & \multirow{2}{*}{CLIP-L~\citep{clip-vit}} & \multirow{2}{*}{Phi3-3.8B~\citep{phi}} & \multirow{2}{*}{558K+665K} & 77.4 & 60.8 & 73.0 & 54.6 & 68.7 & 59.9 & 35.4 & 37.1 \\
        \modelname{} & & & & \textbf{77.6} & \textbf{61.5} & \textbf{74.6} & \textbf{55.2} & \textbf{70.2} & \textbf{61.1} & \textbf{35.8} & \textbf{39.2} \\
        \midrule
        
        LLaVA & \multirow{2}{*}{CLIP-L~\citep{clip-vit}} & \multirow{2}{*}{Mistral-7B~\citep{mistral}} & \multirow{2}{*}{558K+665K} & 79.1 & 62.5 & 72.4 & 56.6 & 70.0 & 63.6 & 36.3 & 47.6 \\
        \modelname{} & & & & \textbf{80.3} & \textbf{64.1} & \textbf{74.5} & \textbf{57.2} & \textbf{72.1} & \textbf{65.1} & \textbf{36.8} & \textbf{48.5} \\
        \midrule

        LLaVA & \multirow{2}{*}{SigLIP-SO~\citep{siglip}} & \multirow{2}{*}{Vicuna-7B~\citep{vicuna}} & \multirow{2}{*}{558K+665K} & 80.8 & 63.2 & 70.6 & \textbf{62.3} & 68.0 & 58.6 & 32.9 & 51.1 \\
        \modelname{} & & & & \textbf{81.2} & \textbf{64.1} & \textbf{71.4} & 62.0 & \textbf{69.6} & \textbf{61.3} & \textbf{34.2} & \textbf{52.3} \\
        \midrule
        
        LLaVA & \multirow{2}{*}{SigLIP-SO~\citep{siglip}} & \multirow{2}{*}{Vicuna-13B~\citep{vicuna}} & \multirow{2}{*}{558K+665K} & 81.8 & 64.3 & 73.8 & \textbf{64.5} & 69.5 & 65.8 & 37.6 & 54.2 \\
        \modelname{} & & & & \textbf{82.4} & \textbf{65.5} & \textbf{75.1} & 64.4 & \textbf{70.8} & \textbf{66.7} & \textbf{38.2} & \textbf{55.3} \\
        
        \bottomrule
    \end{tabular}
    }
    \vspace{-0.2cm}
    \caption{Comparisons between LLaVA~\citep{llava1.5} and \modelname{} when adopting various combinations of vision encoders and LLMs.
    }
    \label{tab:combination}
    \vspace{-0.3cm}
\end{table*}

\noindent  \myparagraph{Analysis of Supervisory Visual Embeddings.}
We utilize refined visual embeddings in the LLM's shallow layers to establish the direct visual supervision.
We analyze three key decision considerations in constructing the supervisory visual embeddings: ($i$) the selection of refined visual embeddings; ($ii$) the integration method of these refined visual embeddings; ($iii$) the degree of supervision on each initial visual embedding. The experiments are conducted using \modelname{}-7B, with the involved LLM Vicuna-7B~\citep{vicuna} consisting of 32 layers. We utilize 10\% of LLaVA-1.5-mix-665k~\citep{llava1.5} in the instruction-tuning stage.

\begin{figure}[t!]
  \centering
  \includegraphics[width=\linewidth]{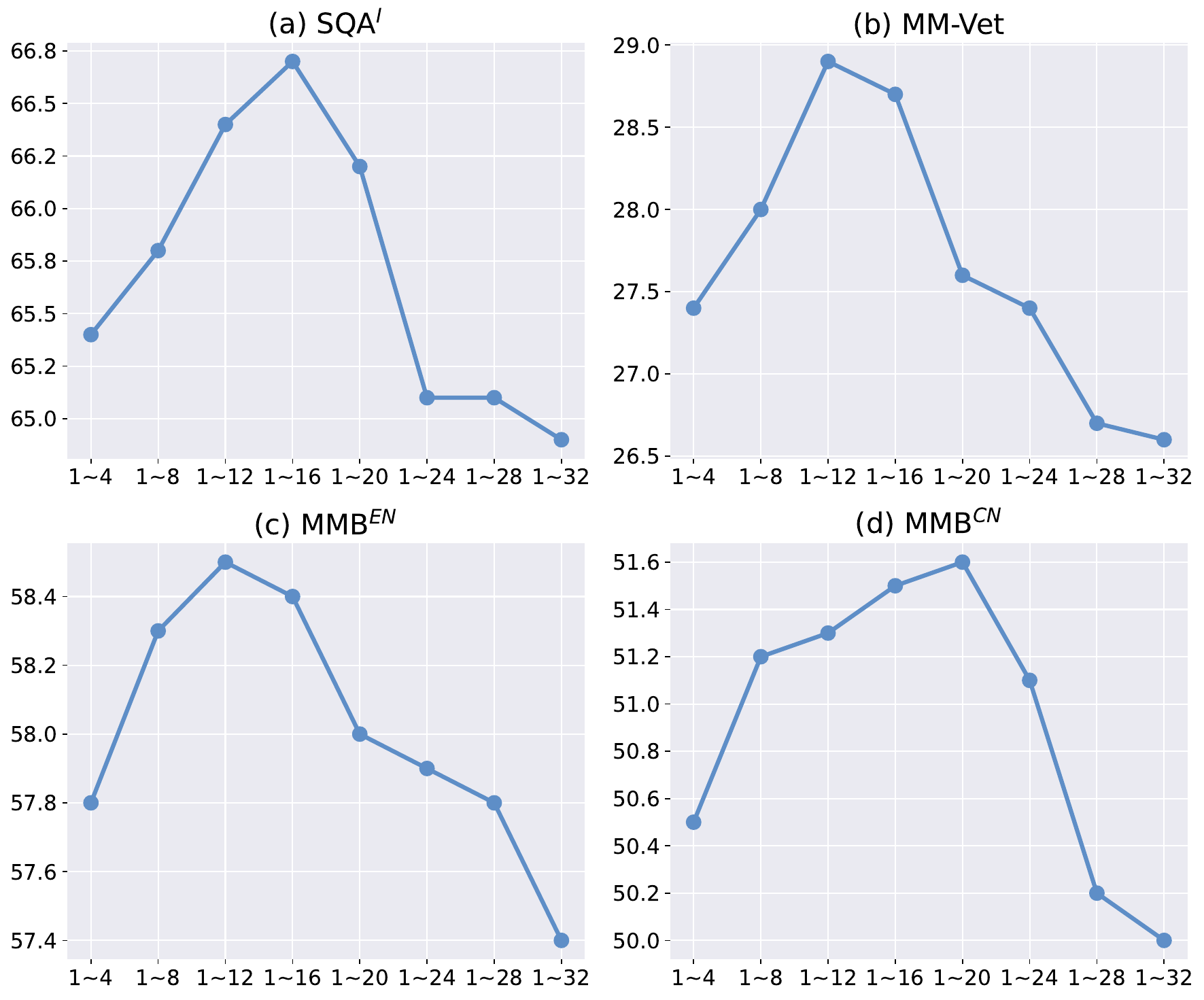}
  \caption{
  The results of utilizing refined visual embeddings from different layers to construct the supervisory visual embedding. The horizontal axis indicates the specific source layers.
  }
  \label{fig:layer_select}
  \vspace{-1.2em}
\end{figure}

\noindent  ($i$) \mysubparagraph{The Selection of Refined Visual Embeddings.}
We explore constructing the supervisory visual embeddings using refined visual embeddings from multiple layers (specifically from the $1\!\sim\!4\text{th}$, $1\!\sim\!8\text{th}$, ..., and $1\!\sim\!32\text{th}$ layers). As illustrated in Figure~\ref{fig:layer_select}, when more refined visual embeddings from the shallow layers (generally $1\!\sim\!16\text{th}$ layers) are adopted, the model performance can gradually improve. However, introducing embeddings from deep layers (generally $16\!\sim\!32\text{th}$ layers) degrades the performance. This implies the pattern difference across the LLM layers. Generally, using refined visual embeddings from the LLM's lower half can construct an appropriate supervision.

\noindent  ($ii$) \mysubparagraph{The Integration of Refined Visual Embeddings.}
In this paper, $w_i$ is used to integrate refined visual embeddings from different layers. 
With utilizing embeddings from the LLM's lower half ($1\!\sim\!l/2$ layers), we explore three settings of the layer weights $w_i$: first, $w_i$ decreases quadratically, $w_i^{\text{dec}}=\frac{(l/2-i + 1)^2}{\sum_{i=1}^{l/2} i^2}$; second, $w_i$ remains constant, $w_i^{\text{const}}=\frac{1}{l/2}$; third, $w_i$ increases quadratically, $w_i^{\text{inc}}=\frac{i^2}{\sum_{i=1}^{l/2} i^2}$. As shown in Table~\ref{tab:lyweights_degree}, $w_i^{\text{inc}}$ maintains better results across various benchmarks. This is largely due to the gradual refinement process within the LLM's shallow layers.

\noindent  ($iii$) \mysubparagraph{The Degree of supervision.}
In the main experiment, we implement fine control of the supervision degree for distinct image patches, which is based on the attention scores from the text to the image. The underlying motivation is that image patches vary in importance and should be treated differently.
We compare it with the scenario where the degree of supervision is equal, namely $a_i$ are all the same.
As shown in Table~\ref{tab:lyweights_degree}, $Sup.^\text{auto}$ achieves better results compared with $Sup.^\text{equal}$, demonstrating the effectiveness of taking the importance of image patches into account.

\begin{table}[t!]
    \small
    \centering
    \begin{tabular}{l|ccc|cc}
        \toprule
        Benchmark & $w_i^\text{dec}$ & $w_i^\text{const}$ & $w_i^\text{enc}$ & $Sup.^\text{equal}$ & $Sup.^\text{auto}$ \\ 
        \midrule
        $\text{VQA}^\text{v2}$ & 72.9 & 73.2 & \textbf{73.4} & 73.0 & \textbf{73.4}\\
        GQA & 55.1 & 55.6 & \textbf{55.9} & 55.6 & \textbf{55.9}\\
        $\text{SQA}^\text{I}$ & 65.6 & 66.0 & \textbf{66.7} & 66.1 & \textbf{66.7}\\
        $\text{MMB}^\text{EN}$ & 57.4 & 57.9 & \textbf{58.4} & 57.6 & \textbf{58.4}\\
        $\text{MMB}^\text{CN}$ & 51.0 & 51.5 & \textbf{51.5} & 51.0 & \textbf{51.5}\\
        MM-Vet & 27.8 & 28.3 & \textbf{28.7} & 27.6 & \textbf{28.7}\\
        
        \bottomrule
    \end{tabular}
    \vspace{-0.1cm}
    \caption{Results under different settings of layers weights and supervision degrees. $w_i^\text{dec}$, $w_i^\text{const}$ and $w_i^\text{enc}$ denote $w_i$ decreasing quadratically, remaining constant, and increasing quadratically respectively. $Sup.^\text{equal}$ and $Sup.^\text{auto}$ denote supervising each image patch equally and based on the importance respectively.
    }
    \label{tab:lyweights_degree}
    \vspace{-0.4cm}
\end{table}

\noindent  \myparagraph{Analysis of Model Settings.}
To evaluate the applicability and robustness of our method, we conduct experiments on a series of MLLMs.
The adopted LLMs include Gemma-2B~\citep{gemma}, Phi3-3.8B~\citep{phi}, Mistral-7B~\citep{mistral}, Vicuna-7B~\citep{vicuna}, and Vicuna-13B~\citep{vicuna}. The vision encoders include CLIP-L~\citep{clip-vit} and SigLIP-SO~\citep{siglip}.
As shown in Table~\ref{tab:combination}, under the combinations of various sizes of LLMs and various resolutions of vision encoders, the introduction of direct visual supervision from the LLM's shallow layers can robustly enhance the model's performance.
\vspace{-4pt}

%% file: sec/6_conclusion.tex
\section{Conclusion}
Modality alignment is a basic issue for MLLMs. The prevailing aligning approach, which solely applies auto-regressive supervision on the texts, has long neglected the introduction of direct supervision for the visual embeddings.
In this work, we conduct a detailed analysis of the visual perception process of MLLMs, revealing the refinement of visual embeddings in the LLM's shallow layers. Based on this, we utilize the refined visual embeddings from the LLM's shallow layers to improve the initial visual embeddings from the perspective of directional alignment and semantic distribution. 
Our method effectively enhances the model's performance, providing valuable insights into the construction of effective direct visual supervision.

%% file: sec/7_acknowledgements.tex
\section*{Acknowledgements}
This research was supported by the National Natural Science Foundation of China (Grant No.62276245).

%% file: sec/X_suppl.tex
\clearpage
\setcounter{page}{1}
\setcounter{section}{0}
\maketitlesupplementary
\renewcommand{\thesection}{\Alph{section}}

\section{\modelname{} Architecture Details}
\label{sec:arch}
In main experiments, the vision encoder adopted in our implementation is CLIP-ViT-L/14-336px~\citep{clip-vit}, which accepts images with a fixed resolution of $336\times336$ pixels and each $14\times14$ sized patch corresponds to an image feature vector. 
The vision projector adopted is a two-layer MLP with GeLU as the activation function. The vision projector converts image features into initial visual embeddings to match the intrinsic dimensions of LLM, and enables LLM to comprehend visual information based on these embeddings. A raw image will produce $576$ visual embeddings.
The LLM adopted is Vicuna-v1.5~\citep{vicuna} which is based on the LLaMA-2 architecture and consists of 7B and 13B parameter versions respectively.

\section{Training Details}
\label{sec:train}
We adopt a two-stage pipeline to train \modelname{}.
In the first stage, we freeze both the vision encoder and the LLM, allowing only the vision projector to be trainable. This stage focuses on achieving preliminary alignment between the visual and text modalities through the vision projector. Training data consists of images and corresponding captions from LLaVA-1.5-558k~\citep{llava1.5}. For the text part, the next-token-prediction loss is applied to all text tokens in the LLM's output space. For the image part, the geometric alignment loss $\mathcal{L}_{das}$ and semantic distribution matching loss $\mathcal{L}_{sds}$ are applied on all initial visual embeddings in the LLM's input space. Specifically, we utilize the weighted summation of refined visual embeddings from $1\!\sim\!16/32$ layers of the Vicuna-v1.5-7B and $1\!\sim\!20/40$ layers of the Vicuna-v1.5-13B as the supervisory visual embedding.

In the second stage, we train both the vision projector and LLM to promote more accurate visual comprehension and enhance the model's instruction-following ability for specific visual tasks. Training data consists of images and corresponding instruction-response pairs from LLaVA-1.5-mix-665k~\citep{llava1.5}. For the text part, the next-token-prediction loss is only applied to the response text tokens. For the image part, the direct visual supervision losses are utilized with the same as the previous stage. In both stages, $\mathcal{L}_{das}$ and $\mathcal{L}_{sds}$ only influence the gradients of the vision projector parameters during backpropagation.
The introduced direct visual supervision does not require additional supervisory models or artificial annotations, making it highly applicable in the training process of a broad range of MLLMs.
The other training hyperparameters are detailed in Table~\ref{tab:hyperparam}.

\begin{table}[t!]
    \small
    \centering
    \resizebox{\linewidth}{!}{
    \begin{tabular}{l|cc}
        \toprule
        \textbf{Hyperparameter} & \textbf{Stage-1} & \textbf{Stage-2} \\ 
        \midrule
        trainable & Vision Projector & Vision Projector/LLM \\
        optimizer & AdamW & AdamW \\
        epoch & 1 & 1 \\
        batch size & 256 & 128 \\
        learning rate & 1e-3 & 2e-5 \\
        warmup ratio & 0.03 & 0.03 \\
        scheduler & cosine & cosine \\
        dtype & bf16 & bf16 \\
        $\lambda_1$ & 1 & 1 \\
        $\lambda_2$ & 0.01 & 0.01 \\
        \bottomrule
    \end{tabular}
    }
    \caption{The training hyperparameters in stage-1 and stage-2.
    }
    \label{tab:hyperparam}
\end{table}



\section{Examples of Visual Perception Process}
\label{sec:mevp}
\begin{figure*}[t!]
  \centering
  \includegraphics[width=\linewidth]{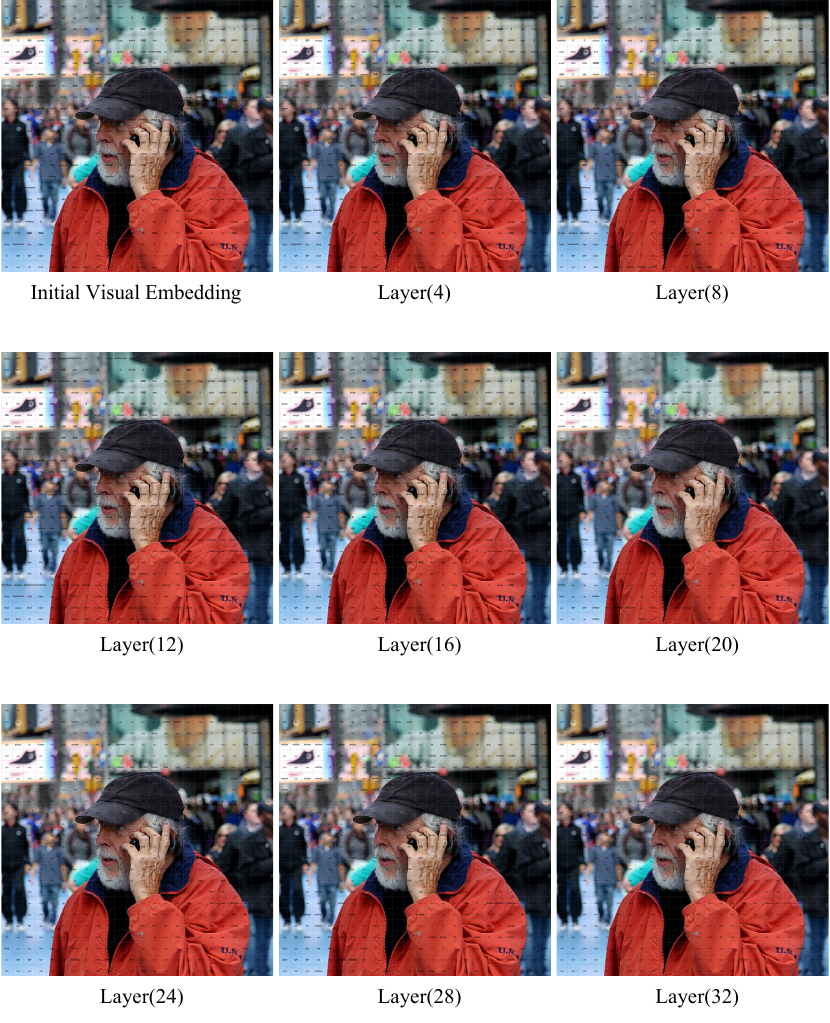}
  \vspace{-0.6cm}
  \caption{
    Visualization of the closest matching token for each visual embedding across the different layers of the LLaVA-1.5~\citep{llava1.5}. The initial visual embeddings are derived from the vision projector and have not yet entered the LLM component.
  }
  \label{fig:app_layer1}
  \vspace{-0.3cm}
\end{figure*}

\begin{figure*}[t!]
  \centering
  \includegraphics[width=\linewidth]{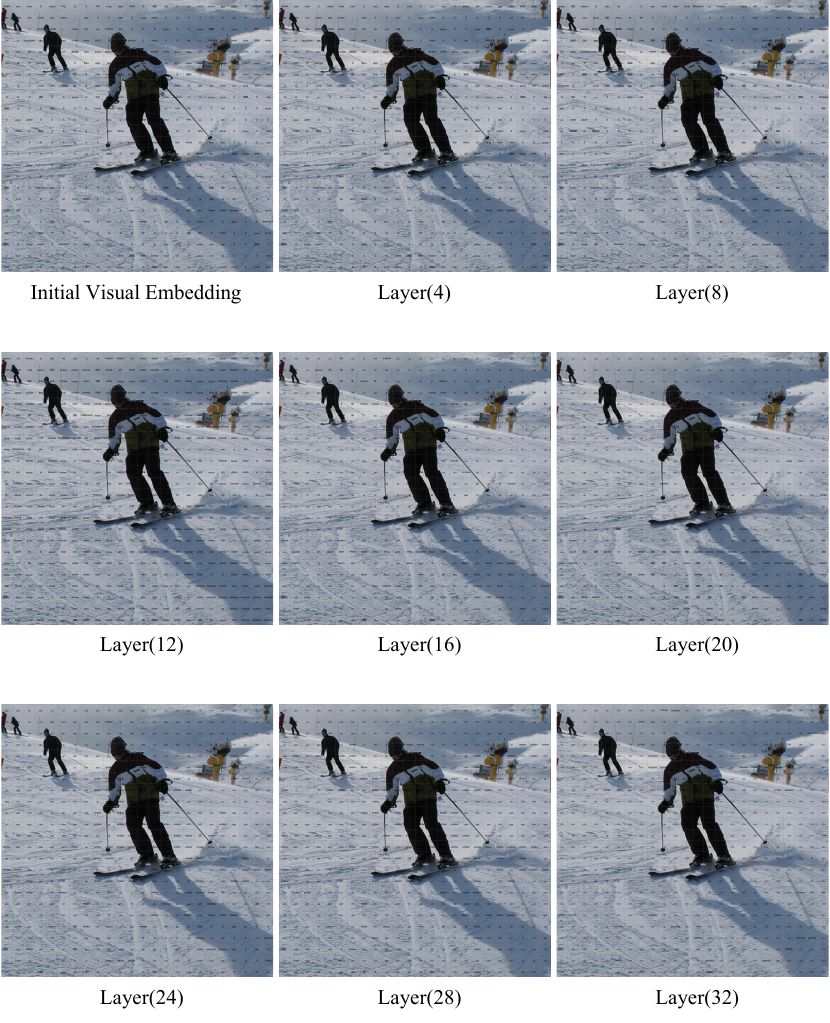}
  \vspace{-0.6cm}
  \caption{
    Visualization of the closest matching token for each visual embedding across the different layers of the LLaVA-1.5~\citep{llava1.5}. The initial visual embeddings are derived from the vision projector and have not yet entered the LLM component.
  }
  \label{fig:app_layer2}
  \vspace{-0.3cm}
\end{figure*}

\begin{figure*}[t!]
  \centering
  \includegraphics[width=\linewidth]{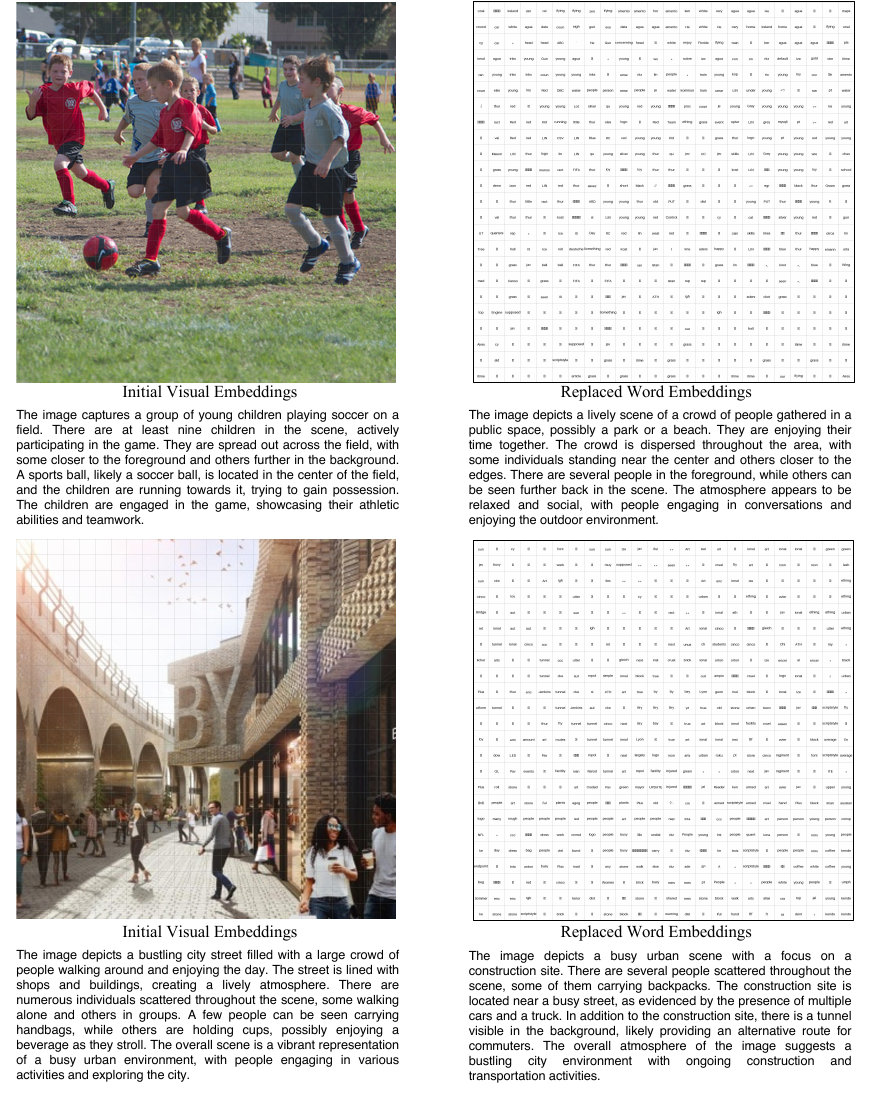}
  \vspace{-0.6cm}
  \caption{
    Visual embeddings from the vision projector are replaced with their closest matching token embeddings in the LLM's vocabulary. LLaVA-1.5~\citep{llava1.5} is then prompted to generate descriptions. The adopted prompt is \textit{Please describe the image in detail}.
  }
  \label{fig:app_replace}
  \vspace{-0.3cm}
\end{figure*}

As illustrated in Figure~\ref{fig:app_layer1} and Figure~\ref{fig:app_layer2}, we provide more examples of the visual perception process within LLaVA-1.5~\citep{llava1.5}. The closest matching token for each visual embedding from different layers of the LLM is obtained based on the cosine similarity. As the vocabulary of LLaVA-1.5 contains tokens for non-English languages as well as special characters, some matching tokens do not display properly. There are significant pattern differences between the LLM’s shallow and deep layers. As illustrated in Figure~\ref{fig:app_replace}, we replace initial visual embeddings with the closest matching token embeddings in the LLM's vocabulary. The LLM is prompted to describe the contents of raw images based on these replaced token embeddings and the adopted prompt is \textit{Please describe the image in detail}. The descriptions are highly consistent with the contents of raw images, which implies that the initial visual embeddings are aligned to the tokens associated with the image patch attributes through multi-modal training and LLMs interprete the text concepts within visual embeddings to understand images. Due to the obvious information loss of replaced token embeddings compared to the initial visual embeddings, the generated descriptions tend to be of lower quality.

\section{Comparisons between LLaVA and BASIC}
\label{sec:compare}
\begin{figure*}[t!]
  \centering
  \includegraphics[width=\linewidth]{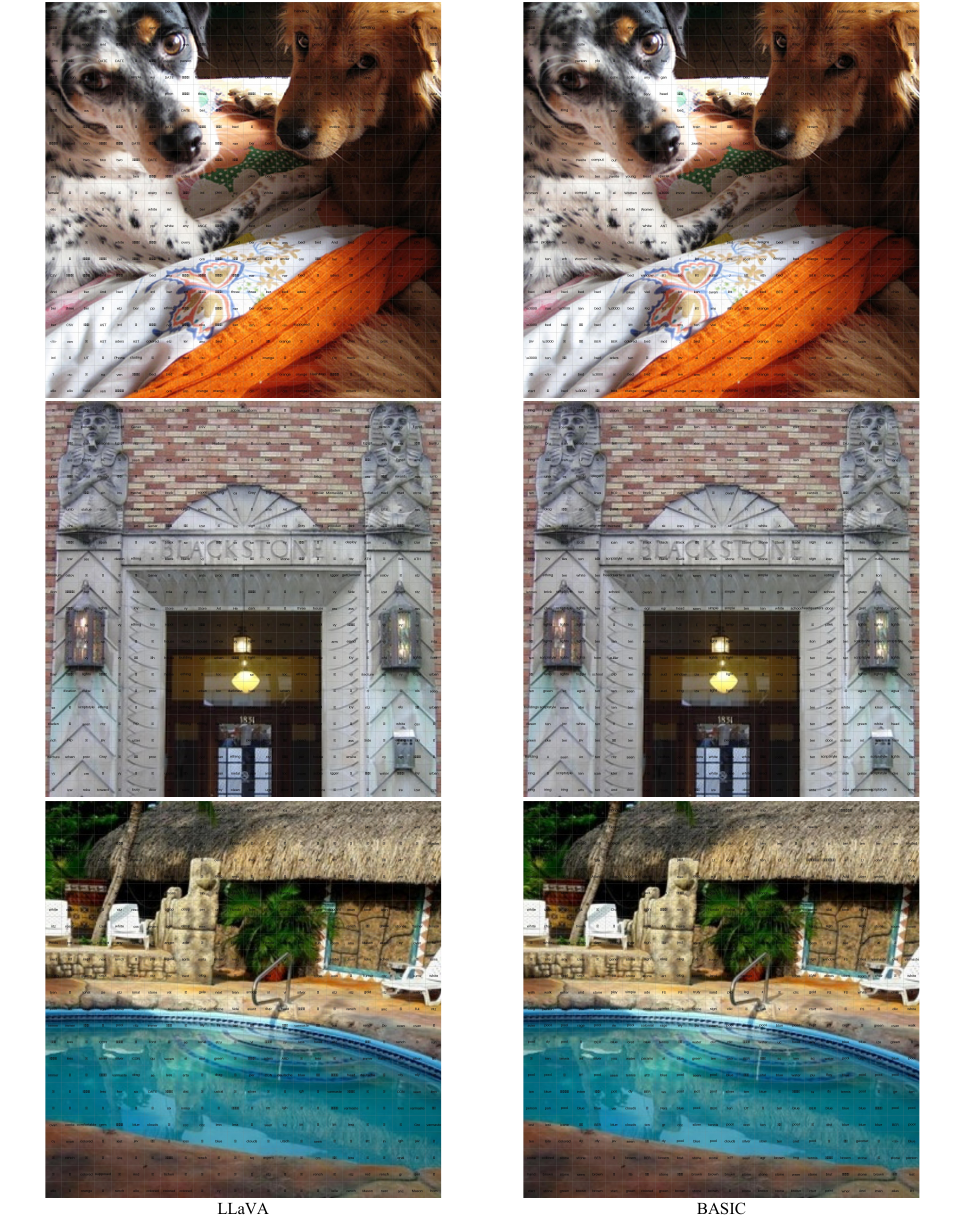}
  \vspace{-0.6cm}
  \caption{
    The respective closest matching token for each initial visual embedding from LLaVA-1.5~\citep{llava1.5} and \modelname{}. Initial visual embeddings from \modelname{} align with more meaningful tokens.
  }
  \label{fig:app_comp}
  \vspace{-0.3cm}
\end{figure*}

As illustrated in Figure~\ref{fig:app_comp}, we visualize the closest matching tokens for initial visual embeddings in LLaVA-1.5~\citep{llava1.5} and \modelname{} respectively. The initial visual embeddings in \modelname{} align with more meaningful tokens. 
As modality alignment plays a basic role in the visual comprehension of MLLMs, \modelname{} demonstrates improved performance across a broad range of benchmarks.

